%% file: arxiv.tex
\documentclass[10pt,twocolumn,letterpaper]{article}

\usepackage{iccv}              
\usepackage{cuted}
\usepackage{capt-of}
\usepackage{multirow}
\usepackage{float} 

\definecolor{iccvblue}{rgb}{0.21,0.49,0.74}
\usepackage[pagebackref,breaklinks,colorlinks,allcolors=iccvblue]{hyperref}

\def\paperID{4047} 
\def\confName{ICCV}
\def\confYear{2025}

\title{FiffDepth: Feed-forward Transformation of Diffusion-Based Generators for Detailed Depth Estimation}

\author{Yunpeng Bai\quad \quad 
Qixing Huang\\
The University of Texas at Austin
}

\begin{document}
\maketitle

\begin{strip}
\centering
\includegraphics[width=0.99\textwidth]{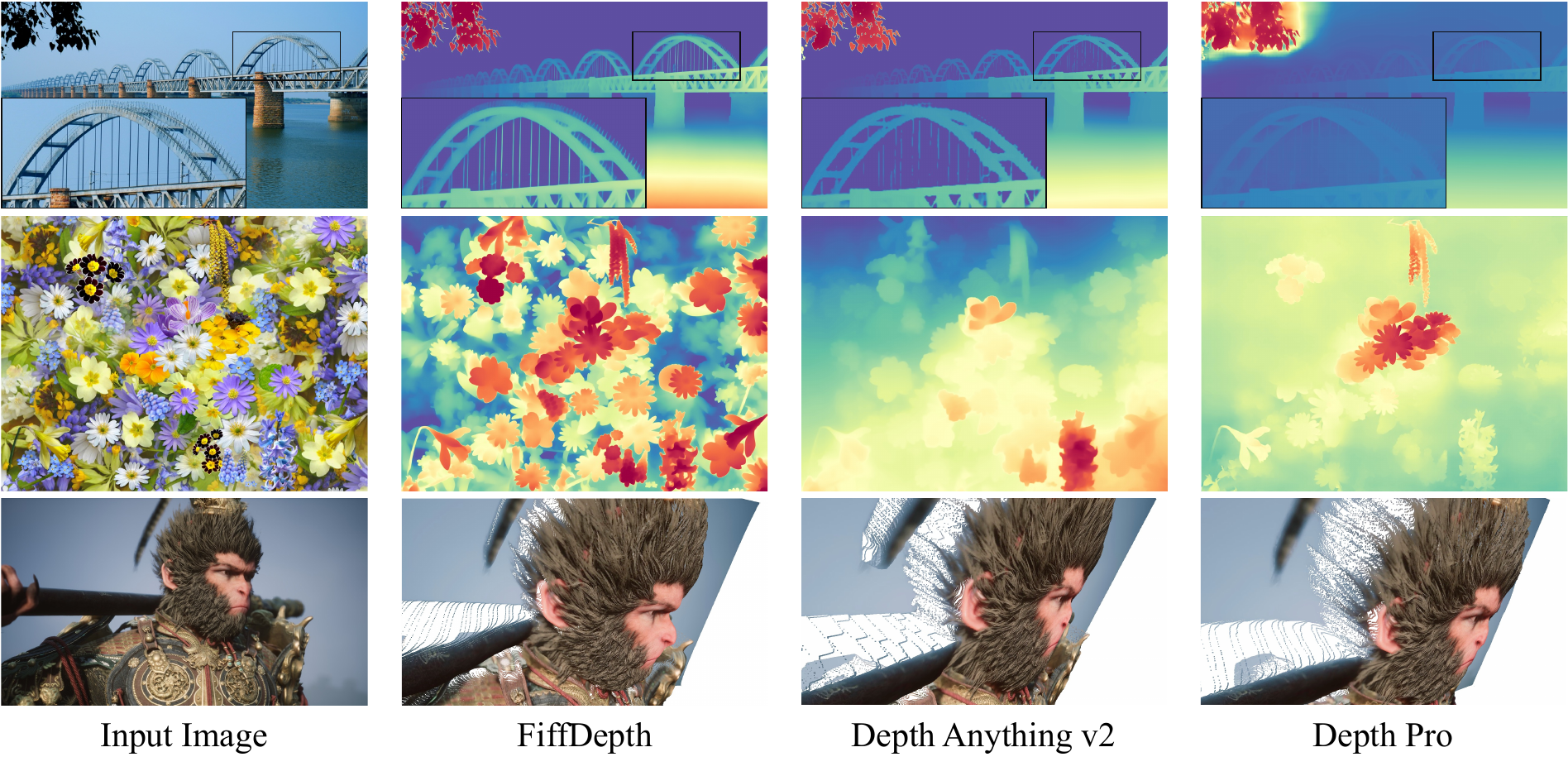}
\vspace{-0.15in}
\captionof{figure}{
Compared to other methods, our model achieves more accurate details and better generalization in depth estimation. The final row shows the point cloud generated from the estimated depth results, and the corresponding depth map can be referenced in Figure \ref{fig:com2}.}
\label{fig:teaser}
\end{strip}

\input{arxiv_sec/0_abstract} 
\input{arxiv_sec/1_intro}
\input{arxiv_sec/2_related}
\input{arxiv_sec/3_method}
\input{arxiv_sec/4_experiments}

\input{arxiv_sec/5_conclusion}

{
    \small
    \bibliographystyle{ieeenat_fullname}
    \bibliography{main}
}

\end{document}

%% file: arxiv_sec/0_abstract.tex
\begin{abstract}

Monocular Depth Estimation (MDE) is a fundamental 3D vision problem with numerous applications such as 3D scene reconstruction, autonomous navigation, and AI content creation. However, robust and generalizable MDE remains challenging due to limited real-world labeled data and distribution gaps between synthetic datasets and real data. Existing methods often struggle with real-world test data with low efficiency, reduced accuracy, and lack of detail. To address these issues, we propose an efficient MDE approach named FiffDepth. The key feature of FiffDepth is its use of diffusion priors. It transforms diffusion-based image generators into a feed-forward architecture for detailed depth estimation. FiffDepth preserves key generative features and integrates the strong generalization capabilities of models like DINOv2. Through benchmark evaluations, we demonstrate that FiffDepth achieves exceptional accuracy, stability, and fine-grained detail, offering significant improvements in MDE performance against state-of-the-art MDE approaches. The paper's source code is available here: \url{https://yunpeng1998.github.io/FiffDepth/}

\end{abstract}

%% file: arxiv_sec/1_intro.tex
\section{Introduction}
\label{sec:intro}

Monocular Depth Estimation (MDE) is a fundamental 3D vision problem with numerous applications in 3D scene reconstruction~\cite{wang2023sparsenerf}, autonomous navigation~\cite{wang2019pseudo}, and more recently in generating AI-based content~\cite{shriram2024realmdreamer}. MDE has made significant progress in the era of deep learning~\cite{ranftl2020towards,yang2024depth2,ke2023repurposing,ranftl2021vision}. Neural networks trained from paired datasets of images and pixel depth exhibit encouraging results and often outperform non-deep learning based counterparts that build on monocular depth cues. Despite significant progress, fundamental challenges remain in efficiency, accuracy, and generalization on diverse in-the-wild data. This is because 1) real depth datasets are usually noisy, and 2) while synthetic data can be used, there exist domain gaps between synthetic datasets and diverse in-the-wild data. 

Specifically, current MDE research relies primarily on synthetic data due to its high-quality annotations and controlled environments. However, the scale and variety of synthetic datasets remain insufficient for comprehensive training. To address this, synthetic-to-real transfer techniques and the utilization of pre-trained models have emerged as viable solutions. Among pre-trained models, generative networks~\cite{rombach2022high} preserve intricate image details more effectively than feed-forward networks (FFNs) like DINOv2~\cite{oquab2023dinov2}, thus holding greater promise for dense prediction models. However, generative models, while detail-rich, often fall short in synthetic-to-real transfer due to their limited generalization capabilities outside the training domain.

Previous studies~\cite{ke2023repurposing,fu2024geowizard} have adopted pre-trained diffusion models, where the main idea is to directly finetuning pre-trained RGB image diffusion models into depth map diffusion generation models conditioned on images. However, this method may not be ideal, as dense prediction models require certainty over diversity. The introduction of any noise or uncertainty during the generation process by these methods is sub-optimal. In contrast, we observe that simply using the denoising diffusion module in a feedforward manner yields better and more stable results. This method capitalizes on extending the trajectories of image diffusion models into the depth domain, representing a significant advancement in both accuracy and efficiency for generative model-based depth estimation methods.

Specifically, we optimize diffusion trajectories for MDE tasks.
To enable the diffusion model to better retain certain detailed generative features when fine-tuned into an MDE model, we preserve the original generative training trajectory while training the model for depth prediction, aiming to maintain the detailed features of the original generative model as much as possible. Furthermore, recognizing the limitations of fintuned diffusion models in maintaining robustness in diverse real-world images—with inaccuracies in predicted depth occurring mainly in the low-frequency components—we leverage the strengths of a DINOv2 \cite{oquab2023dinov2} based model, which excels at predicting accurate low-frequency depth despite its reduced fine detail. To address this issue, we use the diffusion model itself to learn a filter that refines its inaccurate predictions, producing low-frequency outputs with a detail level similar to that of DINOv2’s predictions, thereby matching DINOv2's results and optimizing the low-frequency component of our output. This approach allows us to incorporate a large amount of real image data for training without sacrificing detail preservation of generative models, while simultaneously leveraging DINOv2's strong generalization capabilities to enhance the overall robustness and precision of our MDE models.


In summary, the contributions of our work are as follows: 1) We propose an improved approach for transforming generative models into dense prediction models, specifically for depth prediction tasks, by leveraging diffusion model trajectories in a more stable, feed-forward manner; 2) We introduce a novel distillation method that transfers the robust generalization capabilities of models like DINOv2 to diffusion backbones; 3) Our method demonstrates higher stability, accuracy, and efficiency in depth estimation compared to other approaches based on generative models; 4) Compared to other FFN models, our approach achieves more detailed prediction results, marking a significant advancement in the field of MDE.


%% file: arxiv_sec/2_related.tex
\section{Related Works}
\label{sec:related}

\subsection{Depth Estimation}
Depth estimation has always been a widely researched topic, with numerous studies conducted in the past including single-image \cite{aich2021bidirectional,eigen2014depth,fu2018deep,lee2019big,li2023depthformer,patil2022p3depth,yang2021transformer} and video depth estimation \cite{chen2019self,kopf2021robust,teed2018deepv2d,wang2023neural,yasarla2023mamo}.  Early efforts in image depth estimation, such as DIW \cite{chen2016single} and OASIS \cite{chen2020oasis}, focused on predicting relative (ordinal) depth. 
Subsequent approaches, including MegaDepth~\cite{li2018megadepth} and DiverseDepth~\cite{yin2020diversedepth}, used extensive collections of photographs from the Internet to develop models that adapt to unseen data, while MiDaS~\cite{ranftl2020towards} improved generalization by incorporating a diverse range of datasets during training. Recent advances, including DPT~\cite{ranftl2021vision} and Omnidata~\cite{eftekhar2021omnidata}, adopted transformer-based architectures to improve depth estimation performance. 



However, due to limitations in models and data, these methods exhibit very limited generalization in various open-world scenes. Recently, some efforts \cite{yang2024depth,yang2024depth2,ke2023repurposing,fu2024geowizard} in image depth estimation have made open-world depth estimation feasible by leveraging the power of vast amounts of unlabeled image data and the capabilities of pre-trained generative models.  Although these methods have achieved remarkable progress in the field of depth estimation, challenges such as low efficiency, limited generalization, and insufficient detail preservation remain.


\begin{figure*}
    \centering
    \includegraphics[width=\linewidth]
    {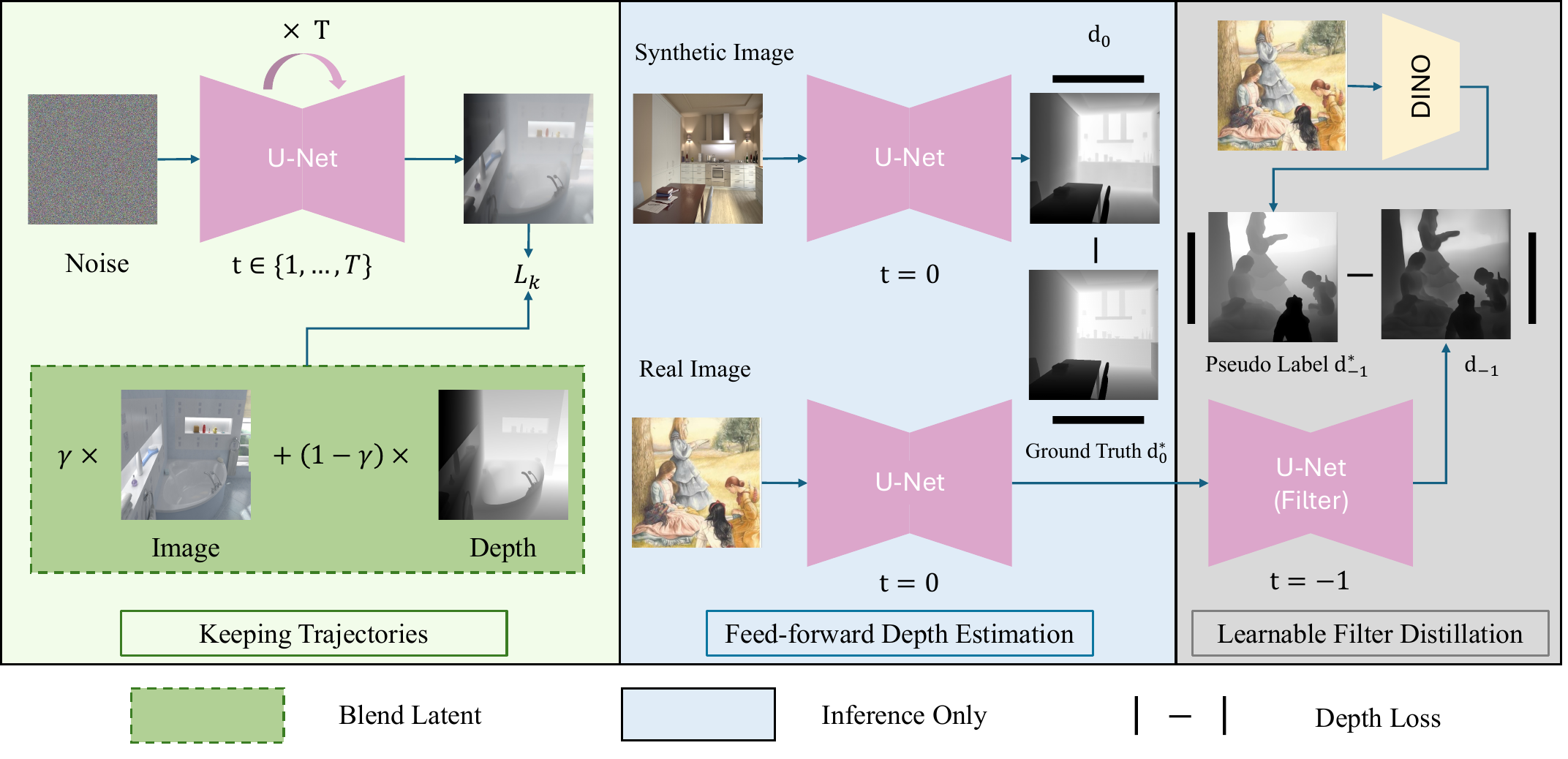}
    \caption{ \textbf{Overview of the proposed method.} To simplify the representation, all the images we used above correspond to the respective latents. We transform the pre-trained diffusion model into a feed-forward approach for depth prediction, using only the result at \( t=0 \) as the output during inference. During training, at \( t=0 \), we use synthetic data to ensure detailed results, while at \( t=-1 \), we leverage pseudo-labels generated by DINOv2 for supervision. }
    \label{fig:pipeline}   
\end{figure*}


\subsection{Diffusion Models as Representation Learner}

The diffusion model training process strongly resembles that of denoising autoencoders (DAE)~\cite{bengio2013generalized,vincent2008extracting,he2022masked}, as both are designed to recover clean images from noise-corrupted inputs. 
 Recent studies have shown that semantic image representations learned from diffusion models can be effectively used for various downstream recognition tasks, such as correspondence~\cite{zhang2024tale}, semantic segmentation~\cite{zhao2023unleashing}, and keypoint detection~\cite{yang2023diffusion}. Notably, the features extracted from diffusion models tend to preserve more intricate details, which has prompted the adoption of pre-trained diffusion models for dense prediction tasks. Recent studies, including Marigold~\cite{ke2023repurposing} and GeoWizard~\cite{fu2024geowizard}, rely on the standard diffusion framework and pre-trained parameters to perform dense prediction tasks. 
Emerging approaches~\cite{xu2024diffusion,he2024lotus,ye2024stablenormal} attempt to bypass the stochastic phase of diffusion models by employing deterministic frameworks.  
However, these adaptations lack deeper exploration of the model's potential, often leading to suboptimal performance and the need for additional post-processing to refine results. 
Moreover, diffusion-based techniques generally exhibit limited generalization capabilities. 
For instance, BetterDepth~\cite{zhang2024betterdepth} incorporates external depth priors as inputs but remains a stochastic framework and heavily depends on the quality of other models. 



%% file: arxiv_sec/3_method.tex
\section{Method}
\label{sec:method}


\subsection{Overview: Feed-forward Transformation of Diffusion Models}
\label{Subsec:Overview}

Fundamentally, generative models construct mappings between a latent space and the ambient data space. These mappings align closely with the needs of depth estimation and other visual recognition tasks, where precise mappings from image data to the corresponding labels are essential. In these tasks, the scarcity of labeled data often limits the precision of the trained models. In the context of learning from a collection of unlabeled data instances, advanced generative models, trained on massive datasets, are capable of learning robust mappings. They exhibit great promise for transferring knowledge to other visual prediction applications~\cite{zhang2024tale,zhao2023unleashing,yang2023diffusion}. Our work builds on this approach, further exploring its application in depth estimation.

Specifically, we finetune a well-trained Stable Diffusion (SD)~\cite{rombach2022high} model to construct our model. A diffusion process constructs multiple intermediate states by progressively adding noise to the data $\mathbf{x}_0$, defined as $
\mathbf{x}_t = \sqrt{\bar{\alpha}_t} \mathbf{x}_0 + \sqrt{1 - \bar{\alpha}_t} \boldsymbol{\epsilon},
$
where 
$\mathbf{\epsilon} \sim \mathcal{N}(\mathbf{0}, \mathbf{I}), \quad \textup{and} \quad  \bar{\alpha}_t:=\prod_{s=1}^t 1-\beta_s
$
with noise schedule $\left\{\beta_1, \ldots, \beta_T\right\}$. Then, diffusion generative model gradually learns the mapping between two distributions by denoising each step, denoted as 
$
x_T \rightarrow x_{T-1} \rightarrow \dots \rightarrow x_0.
$
The diffusion model is primarily a denoising model $\boldsymbol{\epsilon}_\theta$ that follows a loss function 
$
\mathcal{L}_{\text{simple}} = \mathbb{E}_{x_0, \epsilon, t} \left[ \|\boldsymbol{\epsilon} - \boldsymbol{\epsilon}_\theta(x_t, t)\|^2 \right].
$

In depth estimation, the mapping from visual images to depth labels should ideally be deterministic. By leveraging a robust, pre-trained mapping within the generative model, there is no need to decompose the image-label mapping into multiple steps during training. Instead, we can extend the mapping trajectories of the existing diffusion model directly into the depth domain by adapting the learned diffusion process to act as a deterministic one-step feed-forward network. Since our MDE model is built upon the diffusion trajectory for its extension, we set the time step input as \( t = 0 \) in this feed-forward step.
\begin{equation}
\mathbf{d}_0=\hat{\boldsymbol{\epsilon}}_\theta\left(\mathbf{x}_0, t=0\right).
\end{equation}
Since we use diffusion model's parameters to construct our network, we also use $\hat{\boldsymbol{\epsilon}}_\theta$ to represent our network model here. The input $\mathbf{x}_0$ is the latent representation of RGB image, and $\mathbf{d}_0$ is the latent representation of ``depth image," following Marigold's encoding approach. The depth latent $\mathbf{d}_0$ can be reconstructed into a depth map using the VAE of SD with negligible error. While there have been preliminary attempts \cite{xu2024diffusion,he2024lotus} to use similar approaches, they remain in a nascent stage and lack the precision, robustness, and richness in detail needed for effective depth estimation. In the following sections, we will delve into each step of our method, detailing how we adapt the diffusion model to enhance accuracy and detail in depth estimation tasks, such as the key technical contributions of preserving diffusion trajectories and improving synthetic-to-real robustness , as discussed in Section~\ref{Subsec:Keeping:Diffusion} and Section~\ref{Subsec:Improving:Robustness}, respectively.
The overall workflow of the method is illustrated in Figure \ref{fig:pipeline}.

\subsection{Keeping Diffusion Trajectories}
\label{Subsec:Keeping:Diffusion}

Since our approach leverages the trajectory of the diffusion model, it is crucial to prevent degradation of this trajectory during training. 
To achieve this, when fine-tuning the diffusion model to transition it into a feed-forward depth estimator, we simultaneously maintain the feed-forward step along with the preceding denoising training steps from the original diffusion model. While this trajectory was initially developed for image generation, directly applying it as-is does not facilitate an optimal transition to the depth domain. Thus, instead of predicting purely image-based latents, we modify the target latent to be a blend of image and depth representations. 

\begin{equation}
\begin{aligned}
&\mathbf{b}_0 = \gamma \mathbf{x}_0 + (1 - \gamma) \mathbf{d}_0, \\
\mathbf{b}_t=\sqrt{\bar{\alpha}_t}& \mathbf{b}_0+\sqrt{1-\bar{\alpha}_t} \boldsymbol{\epsilon}, t \in\{1, \ldots, T\}.
\end{aligned}
\end{equation}

Here, $\mathbf{b_0}$ represents the blended latent. $\gamma$ controls the balance between the image and depth latents.
In this process, we also use v-prediction re-parameterization approach~\cite{salimans2022progressive} to define the training objective:

\begin{equation}
\begin{gathered}
\mathbf{v}_t=\sqrt{\bar{\alpha}_t} \boldsymbol{\epsilon}-\sqrt{1-\bar{\alpha}_t} \mathbf{b}_0, \\
L_k=\left\|\mathbf{v}_t-\hat{\boldsymbol{\epsilon}}_\theta\left( \mathbf{b}_t, t\right)\right\|_2^2, t \in\{1, \ldots, T\}.
\end{gathered}
\end{equation}
Intuitively, this approach forces the diffusion model to preserve the shared features between the image generation task and the depth estimation task, which are captured in the blended training target. Therefore, it allows the diffusion model to adapt more naturally to depth estimation while retaining essential generative features. Consequently, during fine-tuning, the model maintains features that enhance the accuracy and detail of depth predictions. This part is used exclusively during training. At inference time, our model functions as a fully deterministic framework.

\subsection{Learnable Filter Distillation}
\label{Subsec:Improving:Robustness}

Following previous methods, the above training process only uses synthetic data because it provides high-quality depth Ground Truth. However, this reliance limits both Marigold \cite{ke2023repurposing} and our approach, as SD-based MDE models trained solely on synthetic datasets often struggle to generalize well to in-the-wild data. Nevertheless, in this case, the predictions produced by the SD-based MDE model still retain the necessary details—precisely those details we expect to preserve in the final output. Therefore, enhancing the model's robustness essentially means improving the accuracy of the low-frequency components in the model's output on real image. 

Inspired by prior work \cite{yang2024depth2}, we observe that DINOv2 \cite{oquab2023dinov2}, trained on synthetic data, can generalize effectively to real-world images. However, its depth estimates often lack the necessary fine details—in other words, it can accurately predict the low-frequency depth components for massive real images but misses the high-frequency details. This characteristic aligns well with our needs, so we attempt to leverage the abundance of labels generated by the DINOv2 model to enhance our model's robustness. A straightforward method, as used in Depth Anything v2 \cite{yang2024depth2}, is to use a fintuned DINOv2-G depth model to generate pseudo labels, expanding the training set. However, since DINOv2’s depth predictions lack fine detail, directly incorporating these pseudo labels on the feed-forward output $\mathbf{d}_0$ risks ignoring the detailed features inherent to our model.


To address this, we propose learning a filter, denoted as \(F(d_0)\) that processes our output to remove high-frequency details (e.g., fine details). This filtering produces results at a detail level similar to that of DINOv2, allowing us to focus supervision solely on the less accurate low-frequency components. The filter \(F\) is obtained through learning.  In fact, we observed that the fine-tuned SD model itself already serves as an effective filter learner because the diffusion model is inherently designed to model the subtle differences across different time steps. Therefore, we directly apply an additional SD step on the feed-forward output \(\mathbf{d}_0\) to simulate this filter. To maintain consistency with the inputs and concepts used in the diffusion model, we represent this filter \( F: t_0 \rightarrow t_{-1} \) as the process that transforms the output from \(t_0\) to \(t_{-1}\):





\begin{equation}
\mathbf{d}_{-1}=\hat{\boldsymbol{\epsilon}}_\theta\left(\mathbf{d}_0, t=-1\right).
\label{Eq:d:minusone}
\end{equation}

\begin{figure}
    \centering
    \includegraphics[width=\linewidth]
    {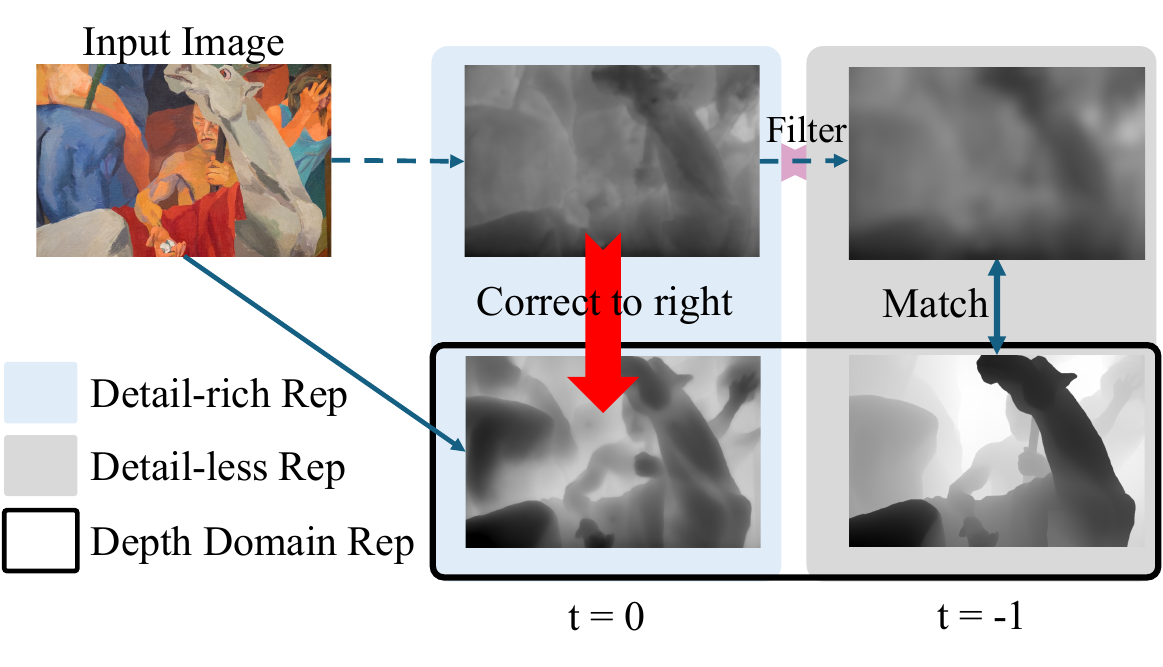}
    \caption{ \textbf{Filter learning.} We use a learnable filter to map our results to detail levels similar to DINOv2’s, matching its outputs and thereby transferring DINOv2’s generalization capabilities to our model without compromising our inherent details.}
    \label{fig:correct}   
\end{figure}

At this stage, we can use the labels predicted by DINOv2 to supervise $\mathbf{d}_{-1}$, allowing us to transfer DINOv2’s robustness without interfering with the detailed features in $\mathbf{d}_0$. In this process, we use real-world image data for \( \mathbf{x}_0 \).  Figure~\ref{fig:correct} illustrates the effects of Eq.~(\ref{Eq:d:minusone}).


\subsection{Final Objective}
\label{Subsec:Final:Objective}

For $\mathbf{d}_0$ and $\mathbf{d}_{-1}$, we follow the method of MiDaS \cite{ranftl2020towards} using MAE loss $L_{\text{MAE}}$ and gradient matching loss $L_{\text{GM}}$ as depth loss. However, unlike the original approach, we apply these losses in the latent space of Stable Diffusion instead:

\begin{equation}
L_{\text{MAE}}\left(\mathbf{d}, \mathbf{d}^*\right) = \frac{1}{M} \sum_{i=1}^{M} |\mathbf{d}_i - \mathbf{d}_i^*|,
\end{equation}
where $\mathbf{d}$ represents the ground truth latent, and $\mathbf{d}^*$ is the model's predicted value. $M$ denotes the
number of pixels in the depth latent.

\begin{equation}
\begin{aligned}
\mathcal{L}_{\text{GM}}\left(\mathbf{d}, \mathbf{d}^*\right)=\frac{1}{M} \sum_{i=1}^M\left(\left|\nabla_x R_i\right|+\left|\nabla_y R_i\right|\right),
\end{aligned}
\end{equation}
where $R_i=\mathbf{d}_i-\mathbf{d}_i^*$. Hence, the final objective is expressed as a weighted sum of the losses $L_{\text{MAE}}$, $L_{\text{GM}}$, and $L_{k}$:

\begin{align}
L_{\text{final}} = \sum_{t \in \{-1, 0\}} & \left( \lambda_{\text{MAE}} L_{\text{MAE}} \left(\mathbf{d}_t, \mathbf{d}_t^*\right) \right. \notag \\
& + \left. \lambda_{\text{GM}} L_{\text{GM}}\left(\mathbf{d}_t, \mathbf{d}_t^*\right) \right) + \lambda_k L_{k}.
\end{align}

$\lambda_{\text{MAE}}$ $\lambda_{\text{GM}}$, and $\lambda_k$ are the weighting factors for their respective loss term. \(\mathbf{d}_0^*\) represents the ground truth (GT) from the synthetic dataset, and \(\mathbf{d}_{-1}^*\) represents the pseudo label generated by DINOv2 for real images.

%% file: arxiv_sec/4_experiments.tex
\section{Experiments}
\label{sec:Experiments}

\begin{table*}
\setlength{\tabcolsep}{2pt}
\begin{tabular}{c|c|cc|cc|cc|cc|cc|c}
\hline \multirow{2}{*}{ \textbf{Method} } & \textbf{Training} & \multicolumn{2}{c}{ \textbf{NYUv2} } & \multicolumn{2}{c}{\textbf{ KITTI }} & \multicolumn{2}{c}{ \textbf{ETH3D} } & \multicolumn{2}{c}{ \textbf{ScanNet} } & \multicolumn{2}{c}{ \textbf{DIODE-Full} } & \multicolumn{1}{c}{\textbf{DA-2K}} \\
& \textbf{Data} & AbsRel $\downarrow$ & $\delta 1 \uparrow$ & AbsRel $\downarrow$ & $\delta 1 \uparrow$ & AbsRel $\downarrow$ & $\delta 1 \uparrow$ & AbsRel $\downarrow$ & $\delta 1 \uparrow$ & AbsRel $\downarrow$ & $\delta 1 \uparrow$ & Acc (\%) \\
 \hline 
DiverseDepth & 320K  & 11.7 & 87.5 & 19.0 & 70.4 & 22.8 & 69.4 & 10.9 & 88.2 & 37.6 & 63.1 & 79.3 \\
MiDaS &2M  & 11.1 & 88.5 & 23.6 & 63.0 & 18.4 & 75.2 & 12.1 & 84.6 & 33.2 & 71.5 & 80.6 \\
LeReS &354K  & 9.0 & 91.6 & 14.9 & 78.4 & 17.1 & 77.7 & 9.1 & 91.7 & 27.1 & 76.6 & 81.1 \\
Omnidata v2 & 12.2M   & 7.4 & 94.5 & 14.9 & 83.5 & 16.6 & 77.8 & 7.5 & 93.6 & 33.9 & 74.2 & 76.8 \\
HDN &300K & 6.9 & 94.8 & 11.5 & 86.7 & 12.1 & 83.3 & 8.0 & 93.9 & 24.6 & 78.0 & 85.7 \\
DPT &1.4M & 9.8 & 90.3 & 10.0 & 90.1 & 7.8 & 94.6 & 8.2 & 93.4 & \textbf{18.2} & 75.8 & 83.2 \\
 Marigold & 74K$^*$  & 5.5 & 96.4 & 9.9 & 91.6 & \underline{6.4} & 96.0 & 6.4 & 95.1 & 30.8 & 77.3 & 86.8 \\
 e2e-ft & 74K$^*$ & 5.2 & 96.6 & 9.6 & 91.9 & \underline{6.4} & 95.9 & 5.8 & 96.2 
 &30.2 & 77.9 & 83.6 \\
 DepthFM & 74K$^*$ & 6.5 & 95.6 & 8.3 & 93.4 & 7.8 & 95.9 & 6.8 & 94.9 & 24.5 & 74.1 & 85.8 \\
 GenPercept & 74K$^*$  & 5.6 & 96.0 & 13.0 & 84.2 & 7.0 & 95.6 & 6.2 & 96.1 & 30.7 & 77.6 & 85.1 \\
Lotus-D & 59K$^*$  & 5.3 & 96.7 & 8.1 & 92.8 & 6.5 & 95.3 & 5.8 & 96.3 & 29.9 & 78.1 & 86.8 \\
Lotus-G & 59K$^*$  & 5.4 & 96.6 & 8.5 & 87.7 & \textbf{6.2} & 96.1 & 6.0 & 96.0 & 29.4 & 78.5 & 86.2 \\
 GeoWizard & 280K$^*$  & 5.2 & 96.6 & 9.7 & 92.1 & \underline{6.4} & \underline{96.1} & $\underline{6.1}$ & 95.3 & 29.7 & \textbf{79.2} & 88.1 \\
DepthAnything v1-L &62.6M$^*$  & \textbf{4.3} & \textbf{98.1} & 7.6 & \textbf{94.7} & 12.7 & 88.2 & \textbf{4.2} & \textbf{98.0} & 27.7 & 75.9 & \underline{88.5} \\
DepthAnything v2-L & 62.6M$^*$ & 4.5 & \underline{97.9} & \underline{7.4} & \underline{94.6} & 13.1 & 86.5 & \textbf{4.2} & 97.8 & 26.2 & 75.4  &\textbf{97.1} \\
FiffDepth (Ours) & 274K$^*$ & \underline{4.4} & 97.8 & \textbf{7.3} & 93.5 & 7.1 & \textbf{97.2} & \textbf{4.2} & \underline{97.9} & \underline{23.9} & \underline{78.1} &\textbf{97.1} \\
\hline
\end{tabular}
\caption{ Quantitative comparison with other affine-invariant depth estimators on several zero-shot benchmarks.  We use AbsRel (absolute relative error: $\left.\left|d^*-d\right| / d\right)$ and $\delta_1$ (percentage of $\max \left(d^* / d, d / d^*\right)<1.25$ ). All metrics are reported as percentages; \textbf{bold numbers} are the best, \underline{underscored} second best. Methods marked with an asterisk (*) utilize pre-trained models.}
\label{tab:depth_comparison}
\end{table*}

\begin{figure*}
    \centering
    \includegraphics[width=\linewidth]
    {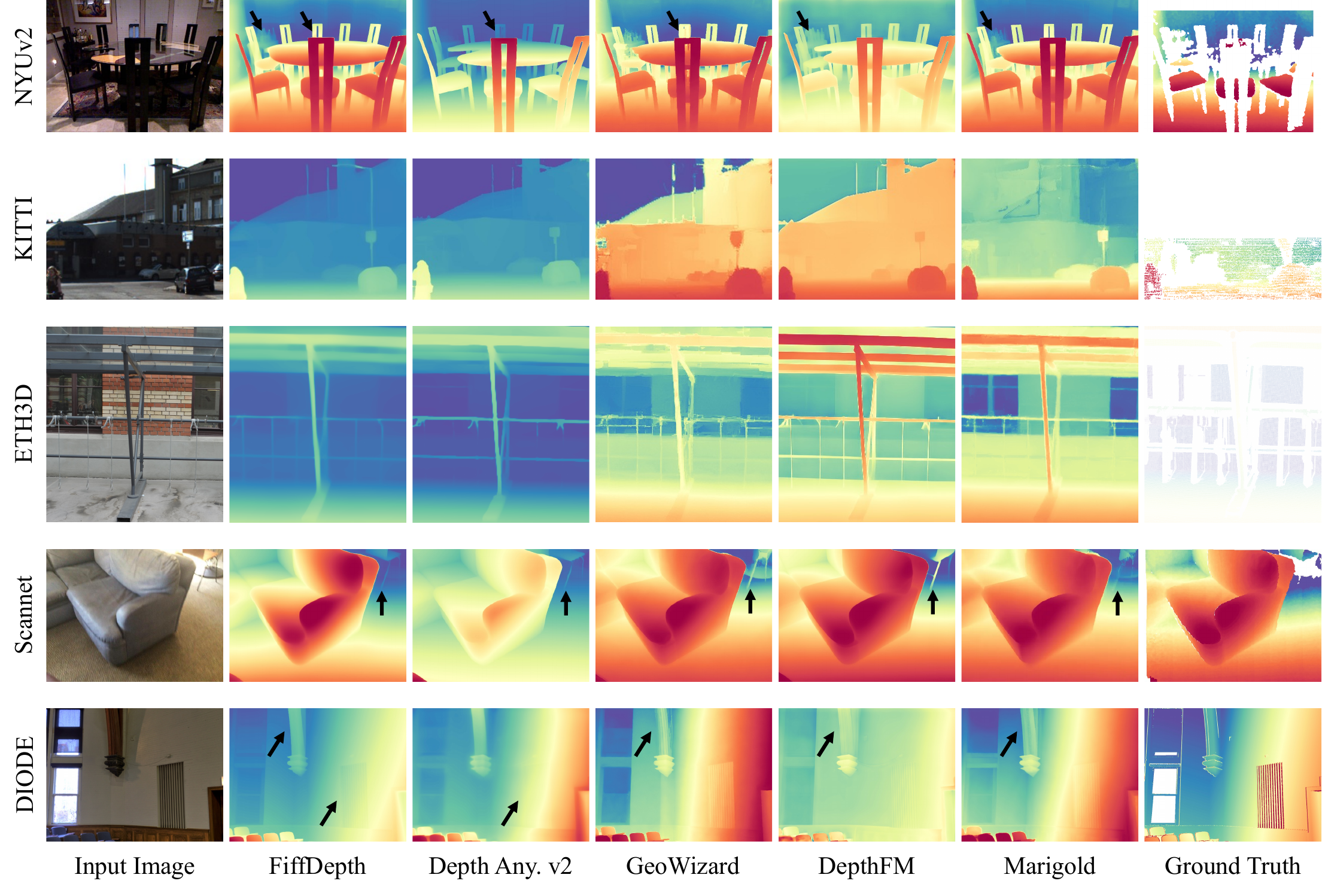}
    \caption{ \textbf{Qualitative comparison across different datasets.} Our method is capable of predicting the depth of various fine objects, such as lampposts, railings, and chair legs.}
    \label{fig:com1}   
\end{figure*}

\begin{figure*}
    \centering
    \includegraphics[width=\linewidth]
    {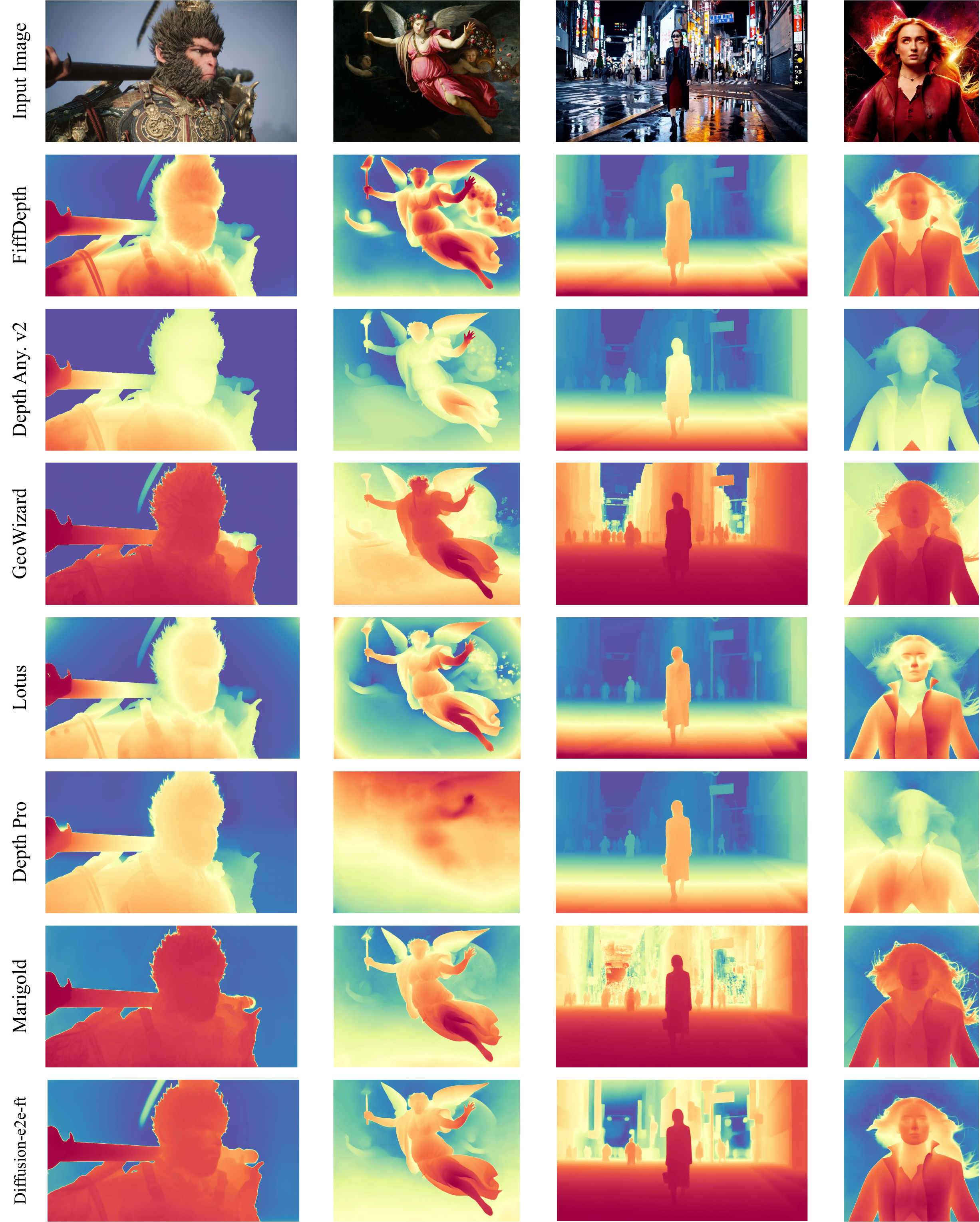}
    \caption{ \textbf{Qualitative comparison on special scenarios.}  
In the special scenarios of games, artworks, AI-generated content, and movies, our method demonstrates strong generalization capability and the ability to predict detailed depth.}
    \label{fig:com2}   
\end{figure*}

\begin{figure*}
    \centering
    \includegraphics[width=\linewidth]
    {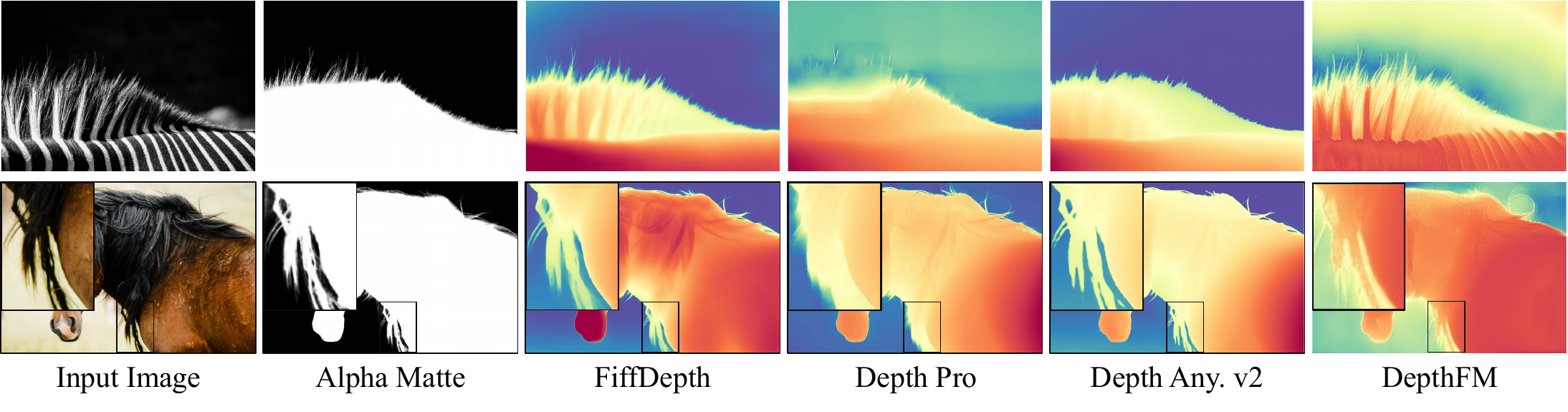}
    \vspace{-0.3in}
    \caption{ \textbf{Boundary visualization comparison.}  These samples are from the AM-2k dataset.}
    \vspace{-0.1in}
    \label{fig:com3}   
    
\end{figure*}

\begin{table*}
\centering
\begin{tabular}{lcccccc}
\hline
\textbf{Method} & \textbf{Sintel F1$\uparrow$} & \textbf{Spring F1$\uparrow$} & \textbf{iBims F1$\uparrow$} & \textbf{AM R$\uparrow$} & \textbf{P3M R$\uparrow$} & \textbf{DIS R$\uparrow$} \\ \hline

DepthAnything v2 & 0.228 & 0.056 & 0.111 & 0.107 & 0.131 & 0.056 \\ 
Depth Pro     & 0.409 & 0.079 & 0.176 & 0.173 & 0.168 & 0.077 \\ 
FiffDepth (Ours)    & \textbf{0.423} & \textbf{0.086} & \textbf{0.189} & \textbf{0.176} & \textbf{0.179} & \textbf{0.091} \\ 
\hline
\end{tabular}
\caption{\textbf{Zero-shot boundary accuracy.} We provide the F1 score for datasets containing ground-truth depth and boundary recall (R) for those with matting or segmentation labels.}
\vspace{-0.1in}
\label{tab:boundary_comparison}
\end{table*}

\begin{table*}[!htbp]
\centering
\begin{tabular}{lccccccc}
\hline
\textbf{Method} & Marigold & Marigold (LCM)  & GeoWizard & DepthFM & DepthAnything v2-L & Depth Pro & Ours \\ \hline
\textbf{Time (s)}      & 103  & 1.7 & 19 & 0.39 & 0.026 & 0.23 & 0.092 \\ 
\hline
\end{tabular}
\caption{\textbf{Running time comparison.} We performe inference on $100$ $512\times512$ images using these methods and report the average time.}
\vspace{-0.1in}
\label{tab:time_comparison}
\end{table*}

\subsection{Implementation Details}
During the training process, we preserve the diffusion trajectory while following the original DDPM noise scheduler \cite{ho2020denoising} using $1000$ diffusion steps.  
To better leverage pre-trained models, we use the Depth Anything V2-Large model as the DINO v2 model for supervision.
This choice was made because DAv2-Giant has not yet released the weights, so we can only use other versions.
Our training dataset consists of two parts. For training at \( t = 0 \) and during trajectory retention, we follow previous approaches and use two synthetic datasets, Hypersim \cite{roberts2021hypersim} and Virtual KITTI \cite{cabon2020virtual}, which cover both indoor and outdoor scenes, with a total of 74K images. For training at $ t = -1 $, we use real-world data from the LAION-Art dataset, a subset of LAION-5B \cite{schuhmann2022laion} containing 8 million samples. However, we observed that training with only 0.2 million samples was sufficient. Synthetic data and real data each account for half of each batch. The parameters are set as follows: \(\gamma = 0.5\), \(\lambda_{\text{MAE}} = 1\), \(\lambda_{\text{GM}} = 0.5\), \(\lambda_{k} = 0.2\).

\subsection{Comparison}
\textbf{Zero-shot affine-invariant depth.} For the evaluation of affine-invariant depth, we use the same datasets and evaluation protocol as Marigold. These datasets include NYUv2 \cite{silberman2012indoor}, ScanNet \cite{dai2017scannet}, KITTI \cite{geiger2012we}, ETH3D \cite{schops2017multi}, and DIODE \cite{vasiljevic2019diode}. We compared FiffDepth with 14 methods that produce affine-invariant depth maps/disparities, all claiming zero-shot generalization capabilities. These include the earlier methods \cite{yin2020diversedepth, yin2021learning, zhang2022hierarchical, ranftl2020towards, ranftl2021vision, eftekhar2021omnidata} , as well as the more recent ones \cite{ke2023repurposing,xu2024diffusion,he2024lotus,martingarcia2024diffusione2eft,gui2024depthfm,fu2024geowizard,yang2024depth,yang2024depth2}.
As shown in Table \ref{tab:depth_comparison}, FiffDepth achieves the best or state-of-the-art comparable results in most test scenarios.  For visualization results, please refer to Figure \ref{fig:com1}. Our method not only accurately predicts the relative depth relationships but also excels in identifying and predicting depth for very fine objects. We also evaluate our method on the DA-2K introduced by Depth Anything v2. On this dataset, our method also performs comparably to Depth Anything v2.

Compared to methods like Depth Anything, which rely on massive training datasets, our model achieves comparable generalization to DAv2 while being trained on only a small amount of real data.
To validate the generalization capability of our method, we present some test examples from special scenarios in Figure \ref{fig:com2}, including games, artworks, AI-generated content, and movies. Our method shows comparable generalization to DAv2 while preserving more details. In contrast, other methods' results are unsatisfactory in both generalization and detail preservation. 

\textbf{Zero-shot boundaries.}
To further demonstrate the accuracy of our method in predicting fine structures, we also employ the Zero-shot Boundaries Metric introduced in the recent work Depth Pro \cite{bochkovskii2024depth} to evaluate boundary sharpness. Following Depth Pro, we compute the depth average boundary F1 score for datasets with ground truth and the boundary recall (R) for datasets with matting or segmentation annotations. The former datasets include Sintel \cite{butler2012naturalistic}, Spring \cite{mehl2023spring}, and iBims \cite{koch2018evaluation}, while the latter include AM-2k \cite{li2022bridging}, P3M-10k \cite{li2021privacy}, and DIS-5k \cite{qin2022highly}. For details on the boundaries metric and its computation, please refer to the Depth Pro paper for further details. Quantitative comparisons in Table \ref{tab:boundary_comparison} demonstrate that our method surpasses Depth Pro and other approaches in boundary prediction. Additionally, the visual results in Figure \ref{fig:com3} further validate that our method predicts more accurate boundaries. 

\begin{figure}
    \centering
    \includegraphics[width=\linewidth]
    {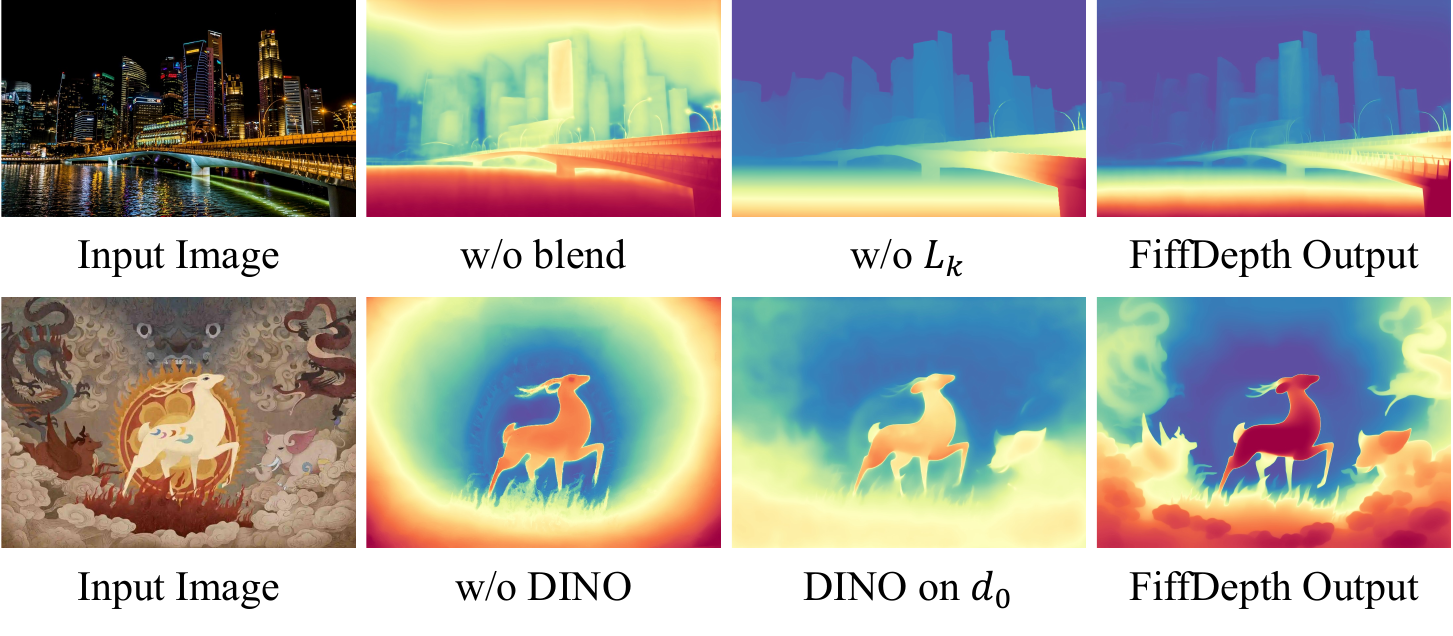}
    \caption{ \textbf{Ablation studies.} The generalization capability and depth details of the method are affected when some essential components are missing.}
    \label{fig:ablation}   
    \vspace{-0.2in}
\end{figure}

\textbf{Running time.} Generative MDE models adopt the diffusion paradigm, and the resulting instability necessitates test-time assembly, leading to a lengthy inference time.

In contrast, our feed-forward approach provides significant efficiency advantages. We evaluate the average inference time for a $512\times512$ image on an NVIDIA Titan RTX GPU. As shown in Table \ref{tab:time_comparison}, our method significantly outperforms other generative approaches in terms of efficiency and achieves performance comparable to DAv2. We tested these methods using their default settings.

\textbf{Ablation studies.}
We conduct ablation studies to validate components of our method. Keeping the diffusion trajectory but predicting purely image latents affects the relative depth relationships between objects (1st row, 1st result in Fig. \ref{fig:ablation}). Without keeping trajectory, some details are lost (1st row, 2nd result in Fig. \ref{fig:ablation}). Omitting DINO supervision impacts the model's generalization ability (2nd row, 1st result in Fig. \ref{fig:ablation}). Using DINO supervision at $d_0$ also reduces details (2nd row, 2nd result in Fig. \ref{fig:ablation}).

%% file: arxiv_sec/5_conclusion.tex
\section{Conclusion}
\label{sec:conclusion}

In this work, we transform diffusion models into stable, feed-forward depth estimators, achieving significant improvements in accuracy and efficiency over generative model-based methods. By combining the detail preservation of generative models with the robust generalization of FFN models like DINOv2, our hybrid approach bridges the synthetic-to-real gap, enhancing stability, predictability, and resolution in MDE for diverse real-world scenarios.
